\documentclass[manuscript,screen]{acmart}
\usepackage[section]{placeins}
\usepackage{multirow}
\usepackage{hyperref}

\AtBeginDocument{%
  \providecommand\BibTeX{{%
    \normalfont B\kern-0.5em{\scshape i\kern-0.25em b}\kern-0.8em\TeX}}}

\setcopyright{acmcopyright}
\copyrightyear{2022}
\acmYear{2022}
\acmDOI{XXXXXXX.XXXXXXX}

\acmConference[Conference acronym 'XX]{Make sure to enter the correct
  conference title from your rights confirmation emai}{June 03--05,
  2018}{Woodstock, NY}
\acmPrice{15.00}
\acmISBN{978-1-4503-XXXX-X/18/06}

\begin{document}

\title{A BERT-based Unsupervised Grammatical Error Correction Framework}


\author{Nankai Lin}
\email{neakail@outlook.com}
\affiliation{%
  \institution{School of Computer Science and Technology, Guangdong University of Technology}
  \city{Guangzhou}
  \state{Guangdong}
  \country{China}
  \postcode{510000}
}

\author{Hongbin Zhang}
\affiliation{%
  \institution{School of Computer Science and Technology, Guangdong University of Technology}
  \city{Guangzhou}
  \state{Guangdong}
  \country{China}
  \postcode{510000}
}

\author{Menglan Shen}
\affiliation{%
  \institution{School of Software and Microelectronics, Peking University}
  \city{Beijing}
  \country{China}
  \postcode{100091}
}

\author{Yu Wang}
\affiliation{%
 \institution{School of South Asian and Southeast Asian Languages and Cultures, Yunnan Minzu University}
 \city{Kunming}
 \state{Yunan}
 \country{China}}

\author{Shengyi Jiang}
\authornote{Shengyi Jiang and Aimin Yang are co-corresponding authors.}

\affiliation{%
  \institution{School of Information Science and Technology, Guangdong University of Foreign Studies}
  \city{Guangzhou}
  \state{Guangdong}
  \country{China}
  \postcode{510000}
}
\email{200511402@oamail.gdufs.edu.cn}

\author{Aimin Yang}
\authornotemark[1]

\affiliation{%
  \institution{School of Computer Science and Technology, Guangdong University of Technology}
  \city{Guangzhou}
  \state{Guangdong}
  \country{China}
  \postcode{510000}
}
\email{amyang@gdut.edu.cn}

\renewcommand{\shortauthors}{Lin and Zhang, et al.}

\begin{abstract}
  Grammatical error correction (GEC) is a challenging task of natural language processing techniques. While more attempts are being made in this approach for universal languages like English or Chinese, relatively little work has been done for low-resource languages for the lack of large annotated corpora. In low-resource languages, the current unsupervised GEC based on language model scoring performs well. However, the pre-trained language model is still to be explored in this context. This study proposes a BERT-based unsupervised GEC framework, where GEC is viewed as multi-class classification task. The framework contains three modules: data flow construction module, sentence perplexity scoring module, and error detecting and correcting module. We propose a novel scoring method for pseudo-perplexity to evaluate a sentence's probable correctness and construct a Tagalog corpus for Tagalog GEC research. It obtains competitive performance on the Tagalog corpus we construct and open-source Indonesian corpus and it demonstrates that our framework is complementary to baseline method for low-resource GEC task.
\end{abstract}




\maketitle

\section{Introduction}
Grammar is a set of specific rules that connects diverse words officially. Grammar errors of writing could hinder readers to understand it. As a remedy to this problem, grammatical error correction (GEC) has become an important research challenge in natural language processing (NLP) research.

Many NLP scholars are interested in GEC in universal languages such as English or Chinese. The low-resource languages, on the other hand, are rarely the topic of research. For low-resource language NLP research, there are very few language resources accessible. This hinders the advancement of low-resource language GEC research and NLP technology.

To solve the lack of a sizable annotated corpus, more GEC of low-resource languages are solved though unsupervised systems, because the seq2seq-based supervised system of GEC is more difficult adapted to low-resource languages than universal languages. Different from the seq2seq-based system, The current investigations adopt the unsupervised technique to generate training data or develop GEC based on the language model scoring method. A language-model-based unsupervised GEC is proposed by Bryant and Briscoe \cite{bryant-briscoe-2018-language}. Compare to the most recent neural and machine translation algorithms that rely on large amounts of annotated data, their basic unsupervised systems also achieve competitive results. Lin et al. \cite{10.1145/3440993} built the first corpus utilized for Indonesian GEC tasks and constructed an unsupervised GEC benchmark supported by neural network models. However, most of the previous unsupervised GEC methods are applied mainly using traditional language model or neural-based approaches, and to the best of our knowledge, pre-trained language model has not been utilized in this sector.

We propose a BERT-based unsupervised GEC framework. The framework regards GEC as the multi-class classification task. It contains three modules: data flow construction module, sentence perplexity scoring module, and error detecting and correcting module. The data flow construction module converts the sentence input to data flow. The sentence perplexity scoring module then scores each converted sentence and uses a scoring mechanism for pseudo-perplexity to evaluate the probable correctness of a sentence. Once errors are found, the sentence is further corrected by the error detection and correction module based on the result of scoring. In addition, we also build a corpus for the research of Tagalog GEC evaluation. We experiment the framework on the Tagalog and open-source Indonesian corpus, which demonstrates that our method achieves better than the state-of-the-art performance. Our framework has the potential to be the new baseline method for low-resource GEC task.

The paper makes the following contributions:

(1) We construct an evaluation corpus for Tagalog GEC. To the best of our knowledge, it is the first GEC corpus for Tagalog.

(2) We propose an unsupervised GEC framework. The framework does not depend on any annotated data.

(3) We propose a scoring method for pseudo-perplexity. Its goal is to efficiently evaluate the likely validity of a sentence.

(4) We conduct experiments on the Tagalog corpus we constructed and open-source Indonesian corpus. Experimental results verify the effectiveness of the proposed model.

\section{Related Work}

The current investigations generate training data using the unsupervised technique and the language model scoring method to develop GEC. However, most of the previous unsupervised GEC methods are applied mainly using traditional language model or neural-based approaches, and to the best of our knowledge, pre-trained language model has not been investigated in this sector. The section focuses on three principal subjects: data synthesis method for unsupervised GEC, language model scoring for unsupervised GEC and sentences scoring based on pre-trained model.

\subsection{Data Synthesis Method for Unsupervised GEC}
In unsupervised grammatical error correction, a common strategy is to perform data synthesis on an unlabeled corpus through an unsupervised method to generate a large amount of pseudo-labeled text. Yasunaga et al. \cite{DBLP:journals/corr/abs-2109-06822} introduced the LM-Critic approach to evaluate sentence grammaticality using a pretrained language model (LM) as a critic. They developed realistic training material from unlabeled text to learn grammatical error correction (GEC) using LM-Critic and the BIFI algorithm. In order to increase the training set size, Solyman et al. \cite{SOLYMAN2021303} developed an unsupervised method to produce synthetic training data on a large scale based on the confusion function. To increase the amount of training data, Grundkiewicz et al. \cite{grundkiewicz-etal-2019-neural} presented a straightforward and unexpectedly efficient unsupervised synthetic mistake generation method based on confusion sets collected from a spellchecker. A Transformer sequence-to-sequence model was pre-trained using synthetic data, which to perform a strong baseline built on real error-annotated data and to make it possible to create a workable GEC system in a situation when real error-annotated data was scarce.

\subsection{Language Model Scoring for Unsupervised GEC}
The LM-scoring-based GEC approach makes the assumption that low probability sentences are more likely to contain grammatical errors than high probability sentences, and the GEC system determines how to transform the former into the latter based on language model probabilities \cite{2211.05166}. Correction candidates can be generated from confusion sets \cite{10.5555/2145432.2145445}, classification-based GEC models \cite{dahlmeier-ng-2012-beam}, or finite-state transducers \cite{stahlberg-etal-2019-neural}.

Bryant and Briscoe \cite{bryant-briscoe-2018-language} proposed a language-model-based approach to unsupervised GEC and investigated if competitive results might be obtained even with extremely basic systems when compared to the most recent neural and machine translation algorithms that rely on large amounts of annotated data. Some hand-crafted rules have been put into GEC systems, and the outcomes were successful. However, manually establishing rules required a lot of time and effort. To address this, Zhang et al. \cite{2206.11569} proposed a technique for automatically mining error templates for GEC. An error template is a regular expression that is used to detect text errors. They have collected 1,119 mistake templates for the Chinese GEC using this methodology. Lichtarge et al. \cite{10.1162/tacl_a_00336} performed an empirical study to discover how to best incorporate delta-log-perplexity, a type of example scoring, into a training schedule for GEC. For Indonesian GEC challenges, Lin et al. \cite{10.1145/3440993} developed an unsupervised GEC framework that might be competitive. This framework handled GEC as a task with multi-classification classifications. In order to fix 10 different types of Part of Speech (POS) errors in Indonesian text, various language embedding models and deep learning models were merged.

\subsection{Sentence Scoring Based on Pre-trained Model}
BERT (Bidirectional Encoder Representation from Transformers) \cite{devlin-etal-2019-bert}, as a non-autoregressive model, is considered unable to evaluate the quality of sentences like language models such as N-grams and RNNLM. However, in recent years, the sentence fluency representation ability of the BERT model has gradually been tapped.

Wang and Cho \cite{wang-cho-2019-bert} showed that BERT is a Markov random field language model and proposed the pseudo-log-likelihood scores (PLLs) to measure the sentence’s scores. Salazar et al. \cite{salazar-etal-2020-masked} evaluated MLMs out of the box via their pseudo-log-likelihood scores (PLLs), which have been computed by masking tokens one by one. They proposed that PLLs outperform scores from autoregressive language models like GPT-2 in a variety of tasks. Furthermore, Futami et al. \cite{wwwww} utilized the pseudo-log-likelihood scores as the evaluation metric in the automatic speech recognition (ASR) task.

\section{Corpus Construction}
We sorted the vocabulary lists corresponding to all parts of speech. From the Tagalog part of speech tagging corpus created by Lin et al. \cite{10.1007/978-3-030-88480-2_17}, we removed parts of speech comprising a significant number of words and parts of speech not applicable to the requirements of Tagalog grammar (error types that should not exist in Tagalog grammar). Finally, as indicated in Table 1, we keep the confusion sets corresponding to 8 types of Tagalog grammatical errors.

We randomly extract sentences containing the confusion set from a large-scale dataset of news texts built by Lin et al. \cite{10.1007/978-3-030-88480-2_17} and intive Tagalog experts to review whether the use of the confusion words contained in these sentences is correct, as well as whether the corresponding types of grammatical errors are accurate. Table 2 demonstrates that the format of the corpus we developed follows that of the Indonesian GEC corpus constructed by Lin et al \cite{10.1145/3440993}. The final Tagalog GEC corpus has 12,953 samples, the statistics for which are shown in Table 3.

\begin{table}
  \caption{Confusion Set of 8 Error Types.}
  \centering
  \label{tab:freq}
  \begin{tabular}{p{3.5cm}p{10.5cm}}
    \toprule
    Error Type & Confusing Word \\
    \hline
Indefinite pronoun & ninuman, anuman, sinumang, sinuman, iba, lahat, ibang, alinman, alinmang, anumang \\
    \hline
Indefinite adverb & saanman, sinuman, kailanman \\
    \hline
Personal pronouns & kami, naming, kita, iyong, ninyo, kaniyang, kanyang, akin, amin, kanya-kanyang, nilang, mong, niya, atin, naming, kang, kaniya, tayo, ka, nya, aming, sila, kanilang, iyo, kayo, ako, ikaw, kanya, nila, sila-sila, kong, kami-kami, natin, siya, kanila, inyo, niyang, nyo, aking, silang, mo, ko \\
    \hline
Preposition & alinsunod, nasa, lampas, sa, patungo, kay, kabilang, tungkol, hanggang, kunti, ng, kina, ukol, kundi, patungong, ayon, bukod, ni, bilang, maliban \\
    \hline
Subordinating conjunction & pag, pagka, kasi, habang, kaya, saka, kahit, hanggang, para, bagama't, pagkat, kapag, na, nung, dahil, maski, nang, tsaka, upang, kung, tuwing, sapagkat, samantala, parang, bilang, bago \\
    \hline
Article & ni, si, ang, ng, nina, sina \\
    \hline
Negative adverb & hindi, wag, huwag, di, hinding-hindi \\
    \hline
Demonstrative & riyan, iyon, ganitong, ganito, iyong, hayaan, ganyan, nandiyan, dun, naroon, gayon, gayun, yung, doon, roon, andito, diyan, nandyan, hayun, rito, ganoon, andyan, yun, yang, iyan, dito, ganoong, yon \\
    
  \bottomrule
\end{tabular}
\end{table}

\begin{table}
  \caption{Examples of Corpus Formats.}
  \centering
  \label{tab:freq}
  \begin{tabular}{p{8.5cm}p{1.3cm}p{2.7cm}}
    \toprule
Sentence & Answer & Error Type \\
\hline
“ Bahala na ang mga residenteng botante sa Makati ang magpasya kung sino ang mas karapat-dapat na umupo sa puwesto , wala akong dapat na piliiin \textbf{[MASK]} sa kanila dahil pareho ko sila anak , ” pahayag ni Binay . & alinman & indefinite pronoun \\
    
  \bottomrule
\end{tabular}
\end{table}

\begin{table}
  \caption{Statistics of Tagalog Grammatical Error Correction Corpus.}
  \centering
  \label{tab:freq}
  \begin{tabular}{cc}
    \toprule
Error Type & Num. of Samples \\
\hline
Indefinite pronoun & 748 \\
Indefinite adverb & 149 \\
Personal pronouns & 3859 \\
Preposition & 1933 \\ 
Subordinating conjunction & 1988 \\
Article & 600 \\
Negative adverb & 493 \\
Demonstrative & 3183 \\
    
  \bottomrule
\end{tabular}
\end{table}

\section{Unsupervised Grammatical Error Correction Framework based on BERT}
We propose a unsupervised framework to correct grammatical error based on the BERT. The framework is composed of three modules: data flow construction module, sentence perplexity scoring module, error detecting and correcting module. First, the data flow construction module converts input sentences into data flow forms. Second, the PS module computes a perplexity score for each of the preceding sentences. The error detecting and correcting module uses each score to determine the input sentence's error and correct it. Finally, the framework identifies and corrects error types in low-resource languages, e.g. eight POS of Tagalog from the third section.

\subsection{Data Flow Construction}
To identify and correct an error type of a token $x_{candidate}$, the data flow construction module needs to construct the data flow $I=\left\{X_1, X_2,\ldots, X_k\right\}$ for a sentence $X=\left\{x_1,x_2,\ldots, x_{candidate},\ldots, x_n\right\}$. First, for the $x_{candidate}$, the data flow construction module utilizes matching way to confirm its POS $P_x$. Then, based on $P_x$, $x_{candidate}$ would be replaced by matching confusing words from the confusion set $C_{p_x}=\left\{c_1, c_2,\ldots, c_k\right\}$. Finally, replaced result could be employed to construct $I$ for the input of the sentence perplexity scoring module.

\subsection{Sentence Perplexity Scoring}
According to $I$, the sentence perplexity scoring module computes the perplexity of every sample through the non-autoregressive pre-trained model. Unlike the pre-trained model, the traditional language model can directly compute the perplexity to represent text fluency, such as N-grams language models \cite{sidorov2014syntactic,damashek1995gauging} and RNNLM models \cite{mikolov2010recurrent,sundermeyer2012lstm}. However, the text perplexity could not be computed directly by the non-autoregressive pre-trained model. Thus, based on the BERT-based pre-trained model, we propose a multi-order perplexity scoring method. Specifically, for a sentence, a novel perplexity scoring mechanism is designed to calculate the first-order perplexity and second-order perplexity separately. Two perplexities are further merged to get the final perplexity score through the weight fusion method.

\subsubsection{First-order perplexity}
For the text first-order perplexity, we calculate it approximately by pseudo perplexity (pseudo-log-likelihood scores, PLLs). The pseudo perplexity computes tokens of $X$ to respective conditional log probability $P_{MLM}(w_t|X_{\textbackslash t})$ and sums all tokens's probabilities. These probabilities are MLM scores, which are obtained through replacing $w_t$ with $[MASK]$. The mathematical expression of the first-order perplexity is as follows:
\begin{equation}\label{eq1}
    PLL_{first}(X):=\sum_{t=1}^{|X|}{\rm log}P_{MLM}(w_t|X_{\textbackslash t})
\end{equation}

Based on the above calculation, we could get the first-order perplexity $s=\left\{s_1, s_2,\ldots,s_k\right\}$ about all samples from $I$. Under the calculation method of the first-order perplexity, the model could measure one masked token to judge whether it can "understand" and "recover" from the remaining sentence information. If the sentence contains incorrect grammatical information (i.e., a grammatical error), the token of other position will be misguided in its prediction, resulting in a higher value of first-order perplexity.

\subsubsection{Second-order perplexity}
We propose the second-order perplexity to address the problem that the first-order perplexity only considers a single token. The model can not only "recover" the masked token after one masked token but can also correctly infer the masked content after multiple masked tokens when more accurate text information is obtained by the model with the firstorder and second-order perplexities. This is because the second-order perplexity applies the more difficult task to require the model to output the scores. To predict masked tokens, the model must "understand" the sentence in which two tokens are masked successively. Compare with the first-order perplexity, since more words are masked and less information is retained by the second-order perplexity, the weight of each retained token would be larger. If the sentence contains incorrect syntactic information, the misdirection of this information will be amplified, and the second-order perplexity will be better able to identify the incorrect sentence. Specifically, We first mask two tokens in succession, but mask successively twice in order when tokens are not in the first or the last position. On the one hand, for the first or the last token, we mask the token only once. The MLM score of the token is directly obtained. On the other hand, according to the twice masking operations, the MLM score of each token is the average of its two MLM scores. Then, based on the above calculated MLM scores of tokens, we can get the second-order perplexity about the sentence. The mathematical expression of the second-order perplexity is as follows:
\begin{equation}\label{eq2}
    PPL_{second}(X):=\sum_{t=1}^{|X|}{\rm log}SOR(t)
\end{equation}
\begin{equation}\label{eq3}
    SOR(t)=\left\{\begin{aligned}
        & P_{MLM}(w_t|X_{ \textbackslash \{0, 1\}}),\quad if\quad t=0 \\
        & \frac {P_{MLM}(w_t|X_{ \textbackslash \{t-1, t\}})+P_{MLM}(w_t|X_{{\textbackslash \{t, t+1\})}}}{2},\quad  if\quad 0<t<|X| \\
        & P_{MLM}(w_{|X|}|X_{ \textbackslash \{|X|-1, |X|\}}),\quad  if\quad t=|X|
            \end{aligned}
\right.
\end{equation}

where SOR($\cdot$) is the MLM score of token about the second-order perplexity.

\subsubsection{Multi-order perplexity}
To obtain the final perplexity score, we design a weight fusion for the first-order and second-order perplexities because they could evaluate the sentence fluency from different levels respectively:
\begin{equation}\label{eq4}
    PPL(X)= \alpha\cdot PLL_{first}(X)+(1-\alpha)\cdot PLL_{second}(X)
\end{equation}

where $\alpha$ is an adjustable weight factor that weighs two perplexities.

\subsection{Error Detecting and Correcting}
For the final perplexity scores $s= \{s_1,s_2,\ldots,s_k \}$ of sentences, we could perform error identification and correction in the EIC. First, $I$ would be sorted by the pseudo perplexity scores. The sentences with smaller scores are ranked higher and the sorted $I$ denoted as $I^\prime=\left\{X_1^\prime,X_2^\prime,\ldots,X_k^\prime\right\}$. Second, corresponding to the sentence with the lowest predicted perplexity score $X_1^\prime$, the confusion word $c_{pred}$ would be matched. Third, the model could determine whether $c_{pred}$ is the same as $x_{candidate}$. If it is the same, the original sentence is correct. If it is different, the model would replace $x_{candidate}$ with $c_{pred}$ to acquire the correct answer.

\section{Experiment and Analysis}
\subsection{Experimental Setup}
To validate with our framework, based on pytorch \footnote[1]{https://github.com/pytorch} with transformers \footnote[2]{https://github.com/huggingface/transformers}, we would run it on RTX 8000 GPUs. It is not only tested on the Tagalog grammar dataset that we created, but it is also validated on the Indonesian grammar dataset created by Lin et al \cite{10.1145/3440993}. We further compare it with the unsupervised grammar correction method called as IndoGEC. That method is applied to Indonesian grammar error correction by Lin et al. \cite{10.1145/3440993}, which is the best unsupervised grammar error correction method based on language model scoring and their internal evaluation method is developed by the RNNLM language model \footnote[3]{https://github.com/kamigaito/rnnlm-pytorch}. Therefore, based on the large-scale news text corpus constructed by Lin et al. \cite{10.1007/978-3-030-88480-2_17}, we construct a ResLSTM-based language model for our framework. The dimension of word vector are 200, the maximum sentence length is 300, the epoch of training is 20 and the learning rate is 5e-5. In addition, for the selection of pre-trained models, we employed two different BERT model for Tagalog \footnote[4]{https://huggingface.co/GKLMIP/bert-tagalog-base-uncased} and Indonesian \footnote[5]{https://huggingface.co/cahya/bert-base-indonesian-1.5G}, respectively.

\subsection{Evaluation Metrics}
Six evaluation metrics are adopted to measure the performance of our model: the accuracy of micro-averaging $P_{micro}$, the accuracy of micro-averaging $R_{micro}$, the F1 value of micro-averaging $F_{{0.5}_{micro}}$, the accuracy of macro-averaging $P_{macro}$, the accuracy of macro-averaging $R_{macro}$, and the F1 value of macro-averaging $F_{{ 0.5}_{macro}}$. It is worth noting that $F_{{0.5}_{macro}}$ is evaluated in two ways, which are averaged $F_{0.5}$ and $F_{0.5}$ of averages. The "averaged $F_{0.5}$" is employed to evaluate our model, but the "$F_{0.5}$ of averages" is utilized by Lin et al in the task of Indonesian grammar error correction. As a result, in order to compare fairly with the method of Lin et al., we report the average F1 values of both macros when evaluating the Indonesian grammar error correction corpus.

\subsection{Experiment in Tagalog Dataset}
\begin{table}
  \caption{Results of comparative experiments}
  \centering
  \label{tab:freq}
  \begin{tabular}{cccccccc}
    \toprule
    
    Method & Type & $P_{macro}$ & $P_{micro}$ & $R_{macro}$ & $R_{micro}$ & $F_{0.5_{macro}}$ & $F_{0.5_{micro}}$  \\
    \hline
    \multirow{9}{*}{IndoGEC}  & Indefinite pronoun & 0.3220  & 0.4225  & 0.3664  & 0.4225  & 0.3185  & 0.4225 \\
 & Indefinite adverb & 0.6052  & 0.7718  & 0.6025  & 0.7718  & 0.5982  & 0.7718 \\
 & Personal pronoun & 0.1711  & 0.1799  & 0.1664  & 0.1799  & 0.1639  & 0.1799 \\
 & Preposition & 0.3791  & 0.3632  & 0.3447  & 0.3632  & 0.3585  & 0.3632 \\
 & Subordinate conjunction & 0.1964  & 0.2007  & 0.1999  & 0.2007  & 0.1873  & 0.2007 \\
 & Article & 0.6639  & 0.6655  & 0.6608  & 0.6655  & 0.6614  & 0.6655 \\
 & Negative adverb & 0.4832  & 0.4260  & 0.4212  & 0.4260  & 0.4256  & 0.4260 \\ 
 & Demonstrative pronoun & 0.1870  & 0.2042  & 0.1871  & 0.2042  & 0.1814  & 0.2042 \\
 & Average & 0.3760  & 0.4042  & 0.3686  & 0.4042  & 0.3618  & 0.4042 \\
    \hline
    \multirow{9}{*}{Our Method}  & Indefinite pronoun & 0.8336  & 0.8008  & 0.7519  & 0.8008  & 0.7693  & 0.8008 \\ 
 & Indefinite adverb & 0.9648  & 0.9732  & 0.9704  & 0.9732  & 0.9657  & 0.9732 \\
 & Personal pronoun & 0.5792  & 0.4628  & 0.4565  & 0.4628  & 0.4942  & 0.4628 \\
 & Preposition & 0.7822  & 0.7977  & 0.7440  & 0.7977  & 0.7592  & 0.7977 \\
 & Subordinate conjunction & 0.6603  & 0.5342  & 0.5316  & 0.5342  & 0.5269  & 0.5342 \\ 
 & Article & 0.9115  & 0.9099  & 0.9084  & 0.9099  & 0.9096  & 0.9099 \\
 & Negative adverb & 0.7207  & 0.6126  & 0.6164  & 0.6126  & 0.6604  & 0.6126 \\ 
 & Demonstrative pronoun & 0.5240  & 0.4907  & 0.4638  & 0.4907  & 0.4604  & 0.4907 \\
 & Average & 0.7470  & 0.6977  & 0.6804  & 0.6977  & 0.6932  & 0.6977 \\
  \bottomrule
\end{tabular}
\end{table}

As shown in Table 4, compared with IndoGEC, our model shows a significant improvement in each part of speech. Among them, from the perspective of the macro-average $F_{0.5}$ value, the performance improvement of the "personal pronoun" type is the most obvious, and $F_{0.5_{macro}}$  has increased by 201.48\%. In addition, the $F_{0.5_{macro}}$ of the “subordinate conjunction” type, “indefinite pronoun” type, “demonstrative pronoun” type and “preposition” type of our method is 181.30\%, 153.80\%, 141.55\% and 111.78\% higher than that of IndoGEC. The IndoGEC model cannot handle types with a large number of confusion words, which is a feature shared by these five types. However, our method makes full use of the large amount of semantic knowledge of the pre-trained model, which can well distinguish the differences between multiple confused words. From the perspective of micro-average $F_{0.5}$ value, the improvement trend of each part of speech is the same as the above analysis. Except for the five parts of speech mentioned above, the $F_{0.5_{micro}}$ of the “indefinite adverb” type has a significant improvement has been significantly improved, which is 61.44\% higher than that of the IndoGEC model. The $F_{0.5_{macro}}$ of our method is 91.58\% higher than IndoGEC on average, and the $F_{0.5_{micro}}$ is 72.61\% higher than IndoGEC on average, which fully demonstrates the superiority of our model. 

\subsection{Ablation Studies}
\begin{table}
  \caption{Results of ablation experiments}
  \centering
  \label{tab:freq}
  \begin{tabular}{cccccccccc}
    \toprule
    Method & Type & $P_{macro}$ & $P_{micro}$ & $R_{macro}$ & $R_{micro}$ & $F_{0.5_{macro}}$ & $F_{0.5_{micro}}$  & $\alpha$\\
    \hline
  \multirow{9}{*}{Our Method} & Indefinite pronoun & 0.8336  & 0.8008  & 0.7519  & 0.8008  & 0.7693  & 0.8008  & 0.9000 \\
 & Indefinite adverb & 0.9648  & 0.9732  & 0.9704  & 0.9732  & 0.9657  & 0.9732  & 0.5000 \\
 & Personal pronoun & 0.5792  & 0.4628  & 0.4565  & 0.4628  & 0.4942  & 0.4628  & 0.3900 \\
 & Preposition & 0.7822  & 0.7977  & 0.7440  & 0.7977  & 0.7592  & 0.7977  & 0.9800 \\
 & Subordinate conjunction & 0.6603  & 0.5342  & 0.5316  & 0.5342  & 0.5269  & 0.5342  & 0.7700 \\ 
 & Article & 0.9115  & 0.9099  & 0.9084  & 0.9099  & 0.9096  & 0.9099  & 0.5000 \\
 & Negative adverb & 0.7207  & 0.6126  & 0.6164  & 0.6126  & 0.6604  & 0.6126  & 0.1200 \\
 & Demonstrative pronoun & 0.5240  & 0.4907  & 0.4638  & 0.4907  & 0.4604  & 0.4907  & 0.8300 \\
 & Average & 0.7470  & 0.6977  & 0.6804  & 0.6977  & 0.6932  & 0.6977  & - \\
 \hline
  \multirow{9}{*}{- First Order PPLs} & Indefinite pronoun & 0.8187  & 0.7580  & 0.7117  & 0.7580  & 0.7313  & 0.7580  & - \\
 & Indefinite adverb & 0.9249  & 0.9530  & 0.9320  & 0.9530  & 0.9263  & 0.9530  & - \\
 & Personal pronoun & 0.5649  & 0.4542  & 0.4589  & 0.4542  & 0.4824  & 0.4542  & - \\
 & Preposition & 0.7708  & 0.7744  & 0.7203  & 0.7744  & 0.7345  & 0.7744  & - \\
 & Subordinate conjunction & 0.5970  & 0.4950  & 0.4947  & 0.4950  & 0.5069  & 0.4950  & - \\
 & Article & 0.9006  & 0.8960  & 0.8949  & 0.8960  & 0.8971  & 0.8960  & - \\
 & Negative adverb & 0.7137  & 0.6085  & 0.6124  & 0.6085  & 0.6550  & 0.6085  & - \\
 & Demonstrative pronoun & 0.5034  & 0.4609  & 0.4365  & 0.4609  & 0.4324  & 0.4609  & - \\
 & Average & 0.7243  & 0.6750  & 0.6577  & 0.6750  & 0.6707  & 0.6750  & - \\
 \hline
 \multirow{9}{*}{- Second Order PPLs} & Indefinite pronoun & 0.8257  & 0.7941  & 0.7448  & 0.7941  & 0.7590  & 0.7941  & - \\
 & Indefinite adverb & 0.9537  & 0.9664  & 0.9671  & 0.9664  & 0.9562  & 0.9664  & - \\
 & Personal pronoun & 0.5941  & 0.4203  & 0.4156  & 0.4203  & 0.4979  & 0.4203  & - \\
 & Preposition & 0.7837  & 0.7982  & 0.7460  & 0.7982  & 0.7620  & 0.7982  & - \\
 & Subordinate conjunction & 0.6485  & 0.5332  & 0.5279  & 0.5332  & 0.5203  & 0.5332  & - \\
 & Article & 0.9099  & 0.9081  & 0.9064  & 0.9081  & 0.9075  & 0.9081  & - \\
 & Negative adverb & 0.7214  & 0.5781  & 0.5825  & 0.5781  & 0.6480  & 0.5781  & - \\
 & Demonstrative pronoun & 0.5124  & 0.4892  & 0.4620  & 0.4892  & 0.4572  & 0.4892  & - \\
 & Average & 0.7437  & 0.6860  & 0.6690  & 0.6860  & 0.6885  & 0.6860  & - \\
  \bottomrule
\end{tabular}
\end{table}

We further conduct ablation experiments on our proposed multi-order pseudo perplexity to explore the effectiveness of our proposed pseudo perplexity calculation method. As shown in Table 5, we can see that after removing the first-order pseudo-perplexity, the performance of all parts of speech declined, the average $F_{0.5_{macro}}$ decreased by 3.26\%, and the average $F_{0.5_{micro}}$ decreased by 3.24\%. What’s more, after removing the second-order pseudo-perplexity, the performance of most parts of speech shows a downward trend, except for the "personal pronoun" type and "preposition" type that have slightly improved. Overall, the average $F_{0.5_{macro}}$ decreased by 1.69\% and the average $F_{0.5_{micro}}$ decreased by 0.68\% when only the first-order pseudo perplexity was retained. It can be seen that the first-order pseudo-perplexity has a greater impact on the model performance than the second-order pseudo-perplexity.

\subsection{Experiments of Recommending Multiple Candidate Words}
\begin{figure}[h]
  \centering
  \includegraphics[width=0.8\textwidth]{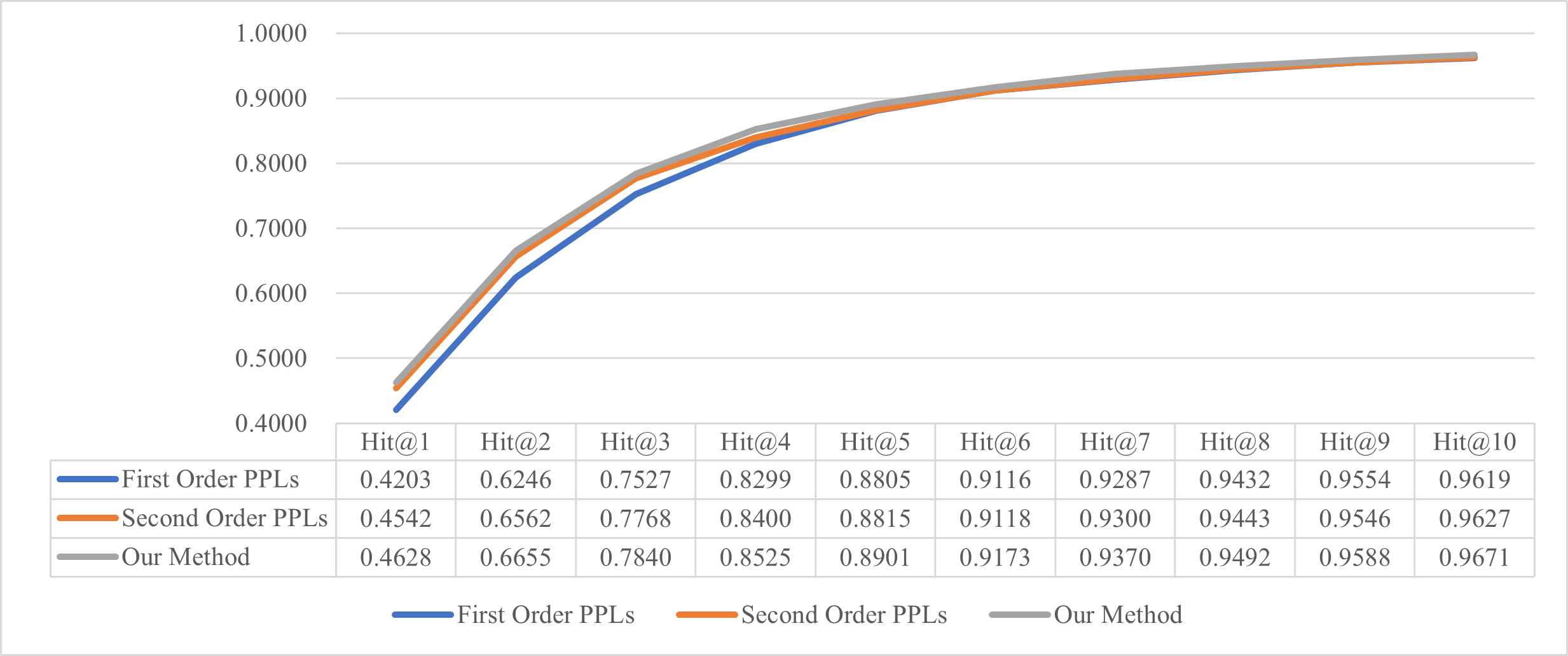}
  \caption{Experimental results of recommending multiple candidate words for personal pronoun.} 
  \label{fig:1} 
\end{figure}

\begin{figure}[h]
  \centering
  \includegraphics[width=0.8\textwidth]{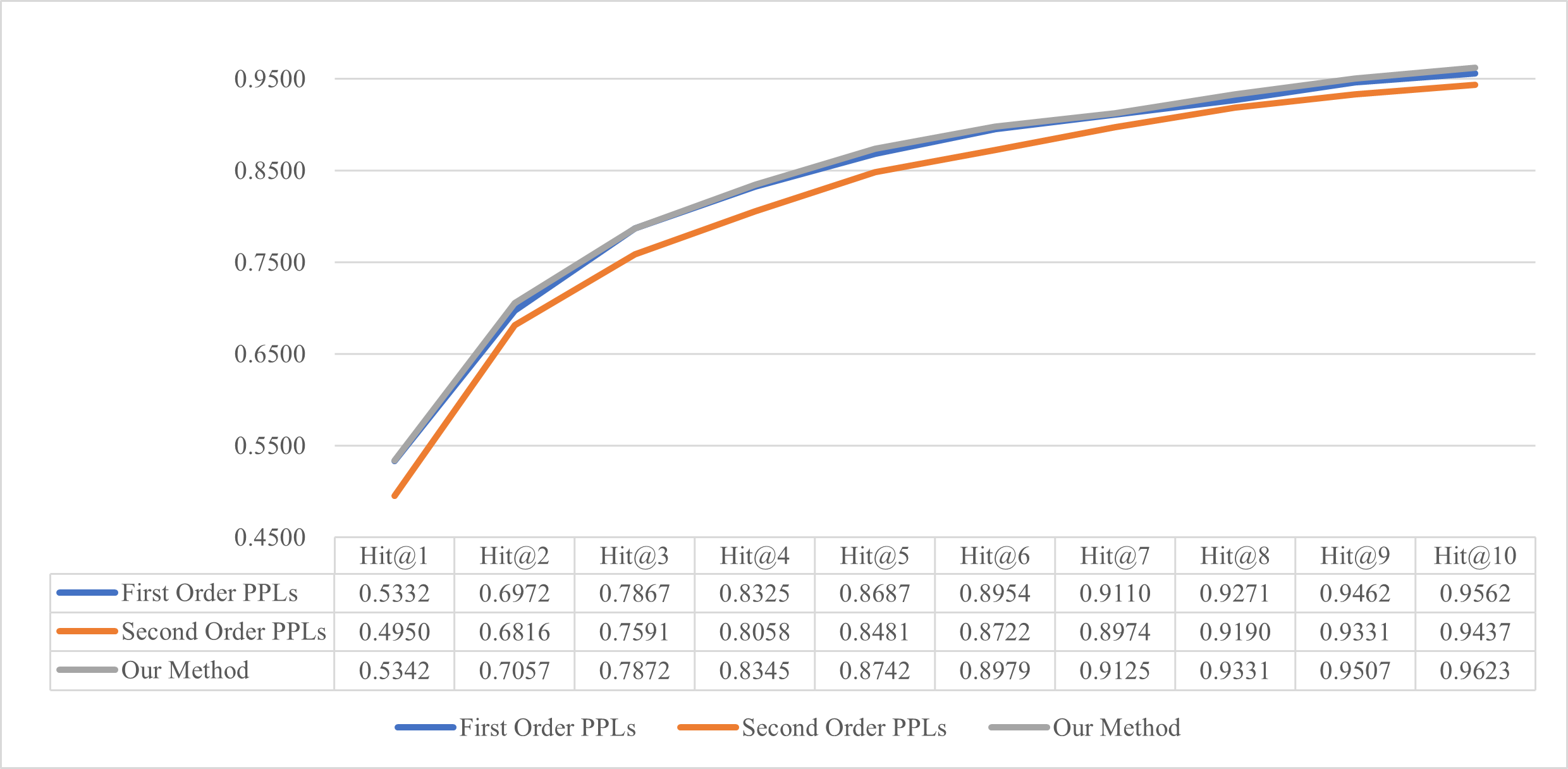}
  \caption{Experimental results of recommending multiple candidate words for subordinate conjunction.} 
  \label{fig:1} 
\end{figure}

\begin{figure}[h]
  \centering
  \includegraphics[width=0.8\textwidth]{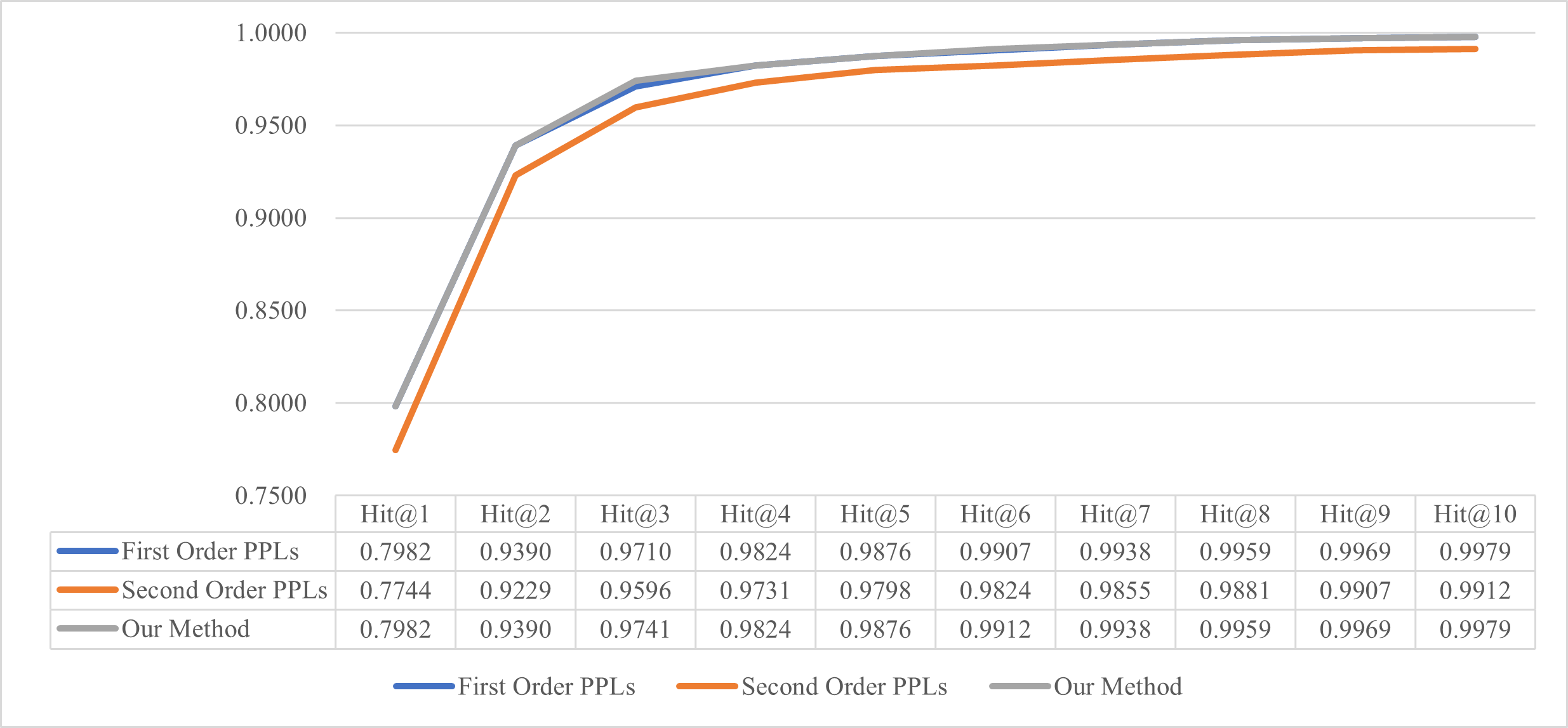}
  \caption{Experimental results of recommending multiple candidate words for preposition.} 
  \label{fig:1} 
\end{figure}

\begin{figure}[h]
  \centering
  \includegraphics[width=0.8\textwidth]{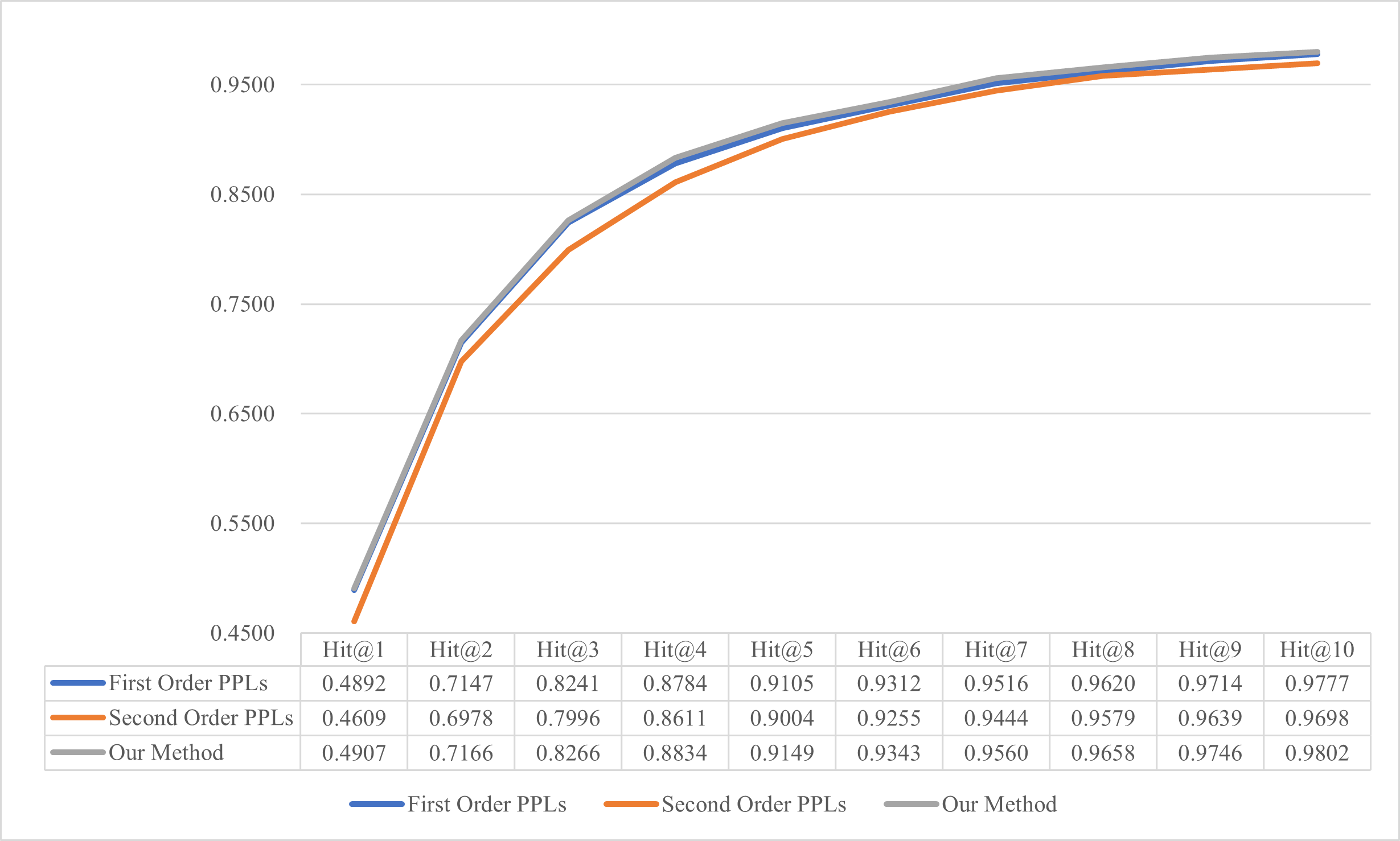}
  \caption{Experimental results of recommending multiple candidate words for demonstrative pronoun.} 
  \label{fig:1} 
\end{figure}

\begin{figure}[h]
  \centering
  \includegraphics[width=0.6\textwidth]{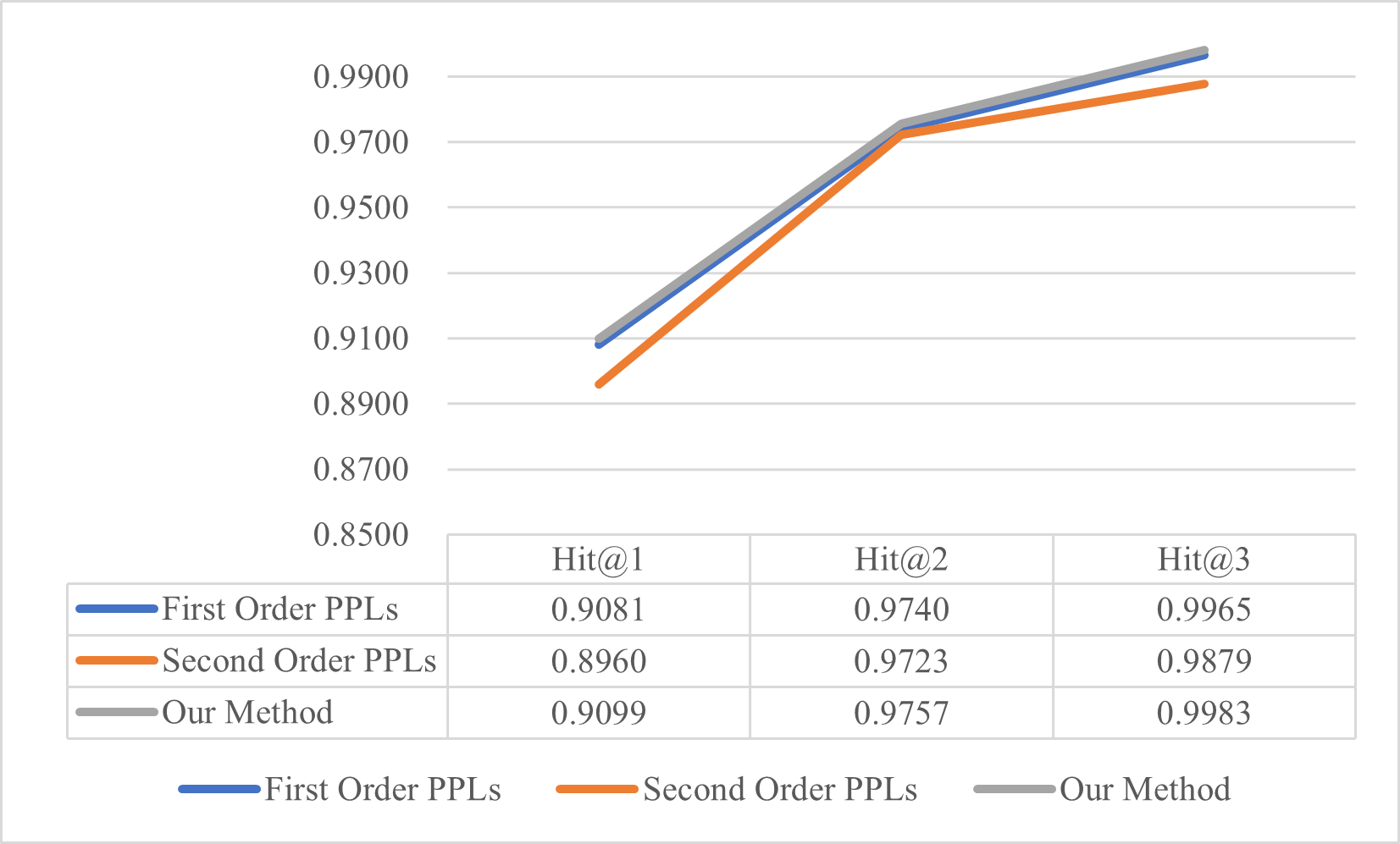}
  \caption{Experimental results of recommending multiple candidate words for article.} 
  \label{fig:1} 
\end{figure}

\begin{figure}[h]
  \centering
  \includegraphics[width=0.6\textwidth]{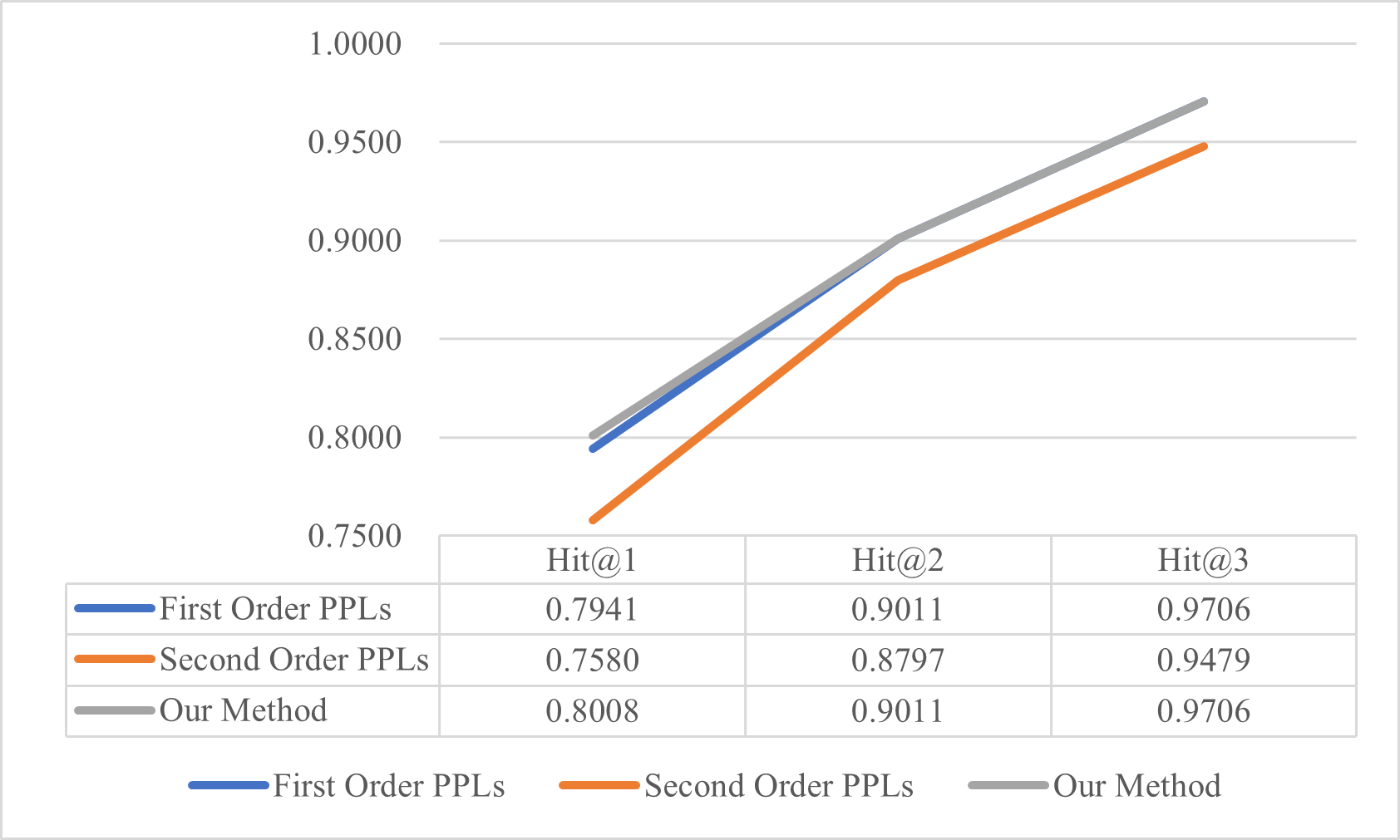}
  \caption{Experimental results of recommending multiple candidate words for indefinite pronoun.} 
  \label{fig:1} 
\end{figure}

\begin{figure}[h]
  \centering
  \includegraphics[width=0.6\textwidth]{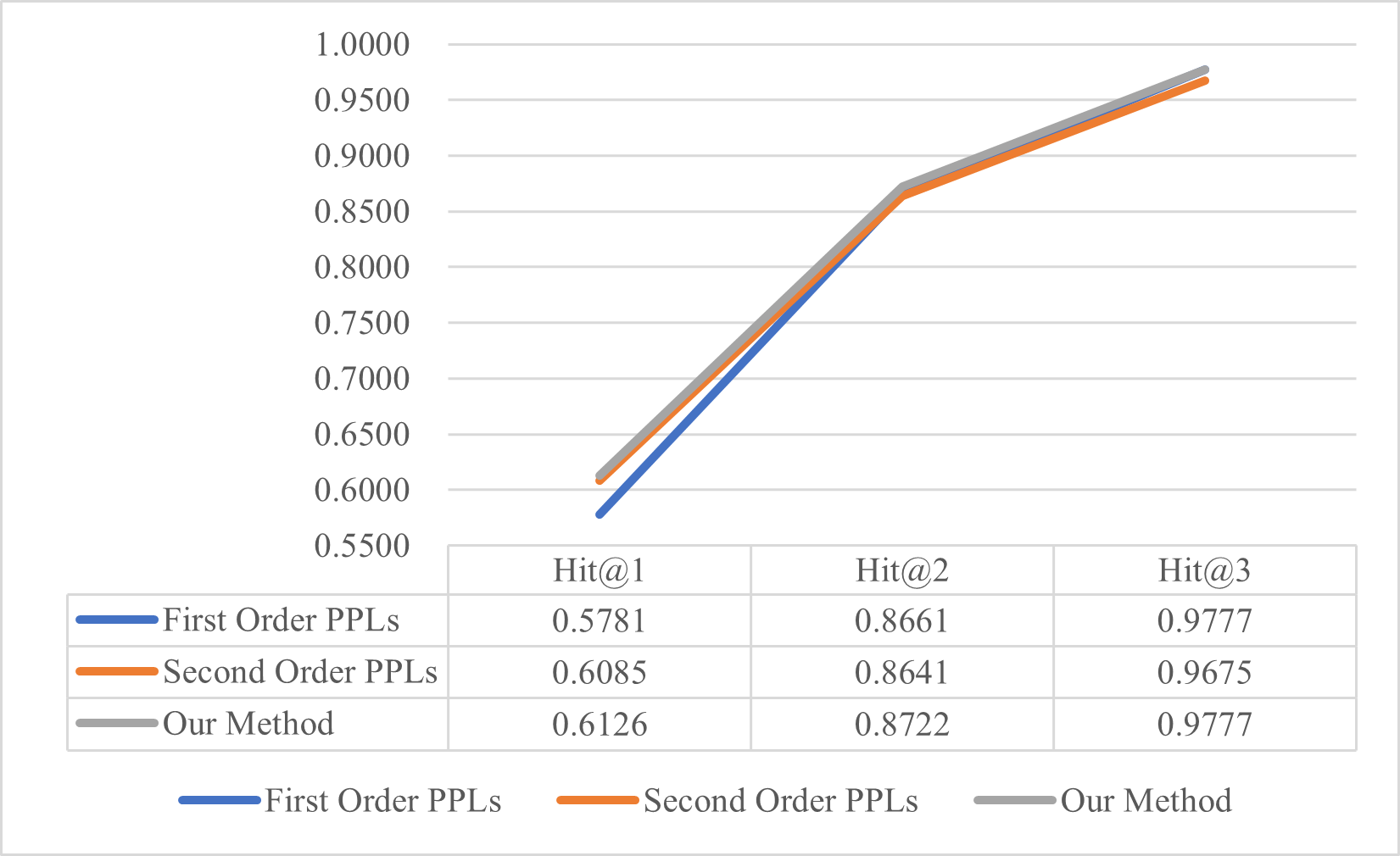}
  \caption{Experimental results of recommending multiple candidate words for negative adverb.} 
  \label{fig:1} 
\end{figure}

Grammatical error correction systems often recommend multiple corrected results as candidate answers in practice. Therefore, we also verified the performance of our model on various part-of-speech types when recommending Top-K results. For part-of-speech types with a large confusion set (the number of confused words contained in the confusion set is greater than 20), our maximum K value is selected as 10. For the part-of-speech type with a small confusion set (the number of confused words contained in the confusion set is less than 10), our maximum K value is 3. In addition, since there are only 3 confusion sets for the "indefinite adverb" type, we do not recommend candidates for this type. The recommendation results of the other seven part-of-speech types are shown in Table 1 to Table 7. In the list of recommended candidate words, the indicator we choose is the accuracy rate Hit@K, that is, whether the recommended candidate words contain the correct answer. It can be seen that the "personal pronoun" type is the best when recommending 8 candidate words (the evaluation index curve area is flat), and the accuracy rate at this time is 94.92\%. The number of best recommendations for “subordinate conjunction”, “preposition”, and “demonstrative pronoun” are 9, 4, and 8, respectively. It can be seen that among the part-of-speech types with a large confusion set, the "subordinate conjunction" type only needs to recommend a small number of words to achieve better performance. When four candidate words are recommended, the accuracy rate is 98.24\%. The other three part-of-speech with smaller confusion sets have the best performance when recommending 3 candidate words, so we believe that the number of best candidate recommendations for these three part-of-speech categories is 3. In addition, it can also be seen that our method performs better than using only the first-order pseudo-perplexity or only the second-order pseudo perplexity when recommending candidate words. However, the three curves (three methods) do not differ much.

\subsection{Experiment in Indonesian Dataset}
\begin{table}
  \caption{Results of comparative experiments for Indonesian}
  \centering
  \label{tab:freq}
  \begin{tabular}{cccccccccc}
    \toprule
    Method & Type & $P_{macro}$ & $P_{micro}$ & $R_{macro}$ & $R_{micro}$ & $F_{0.5_{macro}}$ & $F_{0.5_{micro}}$ & $F_{0.5_{macro}}^{'}$ & $\alpha$\\
    \hline
  \multirow{11}{*}{Second Order PPLs} & Article & 0.8170  & 0.7626  & 0.7626  & 0.7626  & 0.7796  & 0.7626  & 0.8055 & - \\
 & Copula & 0.5431  & 0.6040  & 0.6040  & 0.6040  & 0.5438  & 0.6040  & 0.5543 & -\\
 & Reflexive-pronoun & 0.9798  & 0.9798  & 0.9798  & 0.9798  & 0.9798  & 0.9798  & 0.9798 & -\\ 
 & Relative-pronoun & 0.9234  & 0.9158  & 0.9158  & 0.9158  & 0.9189  & 0.9158  & 0.9219 & -\\
 & Interrogative-pronoun & 0.8678  & 0.8646  & 0.8646  & 0.8646  & 0.8662  & 0.8646  & 0.8672 & -\\
 & Modal-verb & 0.6333  & 0.4343  & 0.4343  & 0.4343  & 0.4851  & 0.4343  & 0.5801 & -\\
 & Demonstrative & 0.6745  & 0.6477  & 0.6477  & 0.6477  & 0.6587  & 0.6477  & 0.6690 & -\\
 & Indefinite Pronoun & 0.6856  & 0.4489  & 0.3673  & 0.4489  & 0.4297  & 0.4489  & 0.5843 & -\\ 
 & Conjunction & 0.3739  & 0.3226  & 0.2582  & 0.3226  & 0.2856  & 0.3226  & 0.3432 & -\\
 & Preposition & 0.5856  & 0.2771  & 0.2636  & 0.2771  & 0.3899  & 0.2771  & 0.4706 & -\\
 & Average & 0.7084  & 0.6258  & 0.6098  & 0.6258  & 0.6337  & 0.6258  & 0.6862 & -\\
 \hline
 \multirow{11}{*}{First Order PPLs} & Article & 0.8212  & 0.8081  & 0.8081  & 0.8081  & 0.8130  & 0.8081  & 0.8185  & - \\
 & Copula & 0.5659  & 0.6182  & 0.6182  & 0.6182  & 0.5647  & 0.6182  & 0.5757  & - \\
 & Reflexive-pronoun & 0.9800  & 0.9798  & 0.9798  & 0.9798  & 0.9799  & 0.9798  & 0.9800  & - \\
 & Relative-pronoun & 0.9086  & 0.8889  & 0.8889  & 0.8889  & 0.8996  & 0.8889  & 0.9046  & - \\
 & Interrogative-pronoun & 0.8871  & 0.8808  & 0.8808  & 0.8808  & 0.8842  & 0.8808  & 0.8858  & - \\
 & Modal-verb & 0.6356  & 0.4687  & 0.4687  & 0.4687  & 0.5225  & 0.4687  & 0.5933  & - \\
 & Demonstrative & 0.6898  & 0.6604  & 0.6604  & 0.6604  & 0.6722  & 0.6604  & 0.6837  & - \\
 & Indefinite Pronoun & 0.6052  & 0.4350  & 0.3467  & 0.4350  & 0.3914  & 0.4350  & 0.5266  & - \\
 & Conjunction & 0.3568  & 0.2069  & 0.1701  & 0.2069  & 0.2248  & 0.2069  & 0.2925  & - \\
 & Preposition & 0.6160  & 0.2853  & 0.2763  & 0.2853  & 0.3993  & 0.2853  & 0.4945  & - \\
 & Average & 0.7066  & 0.6232  & 0.6098  & 0.6232  & 0.6352  & 0.6232  & 0.6849  & - \\
 \hline
 \multirow{11}{*}{Our Method} & Article & 0.8212  & 0.8081  & 0.8081  & 0.8081  & 0.8130  & 0.8081  & 0.8185  & 0.9900 \\
 & Copula & 0.5729  & 0.6323  & 0.6323  & 0.6323  & 0.5733  & 0.6323  & 0.5839  & 0.6200 \\
 & Reflexive-pronoun & 0.9800  & 0.9798  & 0.9798  & 0.9798  & 0.9799  & 0.9798  & 0.9800  & 0.7100 \\
 & Relative-pronoun & 0.9302  & 0.9226  & 0.9226  & 0.9226  & 0.9263  & 0.9226  & 0.9287  & 0.7200 \\
 & Interrogative-pronoun & 0.8907  & 0.8828  & 0.8828  & 0.8828  & 0.8871  & 0.8828  & 0.8891  & 0.7800 \\
 & Modal-verb & 0.6362  & 0.4707  & 0.4707  & 0.4707  & 0.5241  & 0.4707  & 0.5944  & 0.9900 \\
 & Demonstrative & 0.7057  & 0.6806  & 0.6806  & 0.6806  & 0.6910  & 0.6806  & 0.7005  & 0.7900 \\ 
 & Indefinite Pronoun & 0.6858  & 0.4506  & 0.3749  & 0.4506  & 0.4317  & 0.4506  & 0.5882  & 0.1100 \\ 
 & Conjunction & 0.3843  & 0.3226  & 0.2581  & 0.3226  & 0.2874  & 0.3226  & 0.3501  & 0.0700 \\
 & Preposition & 0.6164  & 0.2896  & 0.2797  & 0.2896  & 0.4038  & 0.2896  & 0.4968  & 0.9300 \\
 & Average & 0.7223  & 0.6440  & 0.6289  & 0.6440  & 0.6518  & 0.6440  & 0.7015  & - \\
 \hline
  \multirow{11}{*}{IndoGEC} & Article & 0.5000  & - & 0.5000  & - & - & - & 0.5000  & - \\
 & Copula & 0.4440  & - & 0.4670  & - & - & - & 0.4480  & - \\
 & Reflexive-pronoun & 0.8230  & - & 0.8180  & - & - & - & 0.8220  & - \\
 & Relative-pronoun & 0.7660  & - & 0.5020  & - & - & - & 0.6930  & - \\
 & Interrogative-pronoun & 0.6760  & - & 0.6120  & - & - & - & 0.6620  & - \\
 & Modal-verb & 0.4440  & - & 0.2570  & - & - & - & 0.3880  & - \\
 & Demonstrative & 0.5800  & - & 0.3190  & - & - & - & 0.4980  & - \\
 & Indefinite Pronoun & 0.4710  & - & 0.2230  & - & - & - & 0.3850  & - \\
 & Conjunction & 0.5960  & - & 0.2900  & - & - & - & 0.4920  & - \\
 & Preposition & 0.6380  & - & 0.2730  & - & - & - & 0.5030  & - \\
 & Average & 0.5938  & - & 0.4261  & - & - & - & 0.5510  & - \\
  \bottomrule
\end{tabular}
\end{table}

We further carry out experiments on the public Indonesian grammatical error correction corpus, and the experimental results are shown in Table 6. Our method is 20.05\% higher in $F_{0.5_{macro}}^{'}$ value than the SOTA method on this dataset, reflecting the effectiveness of our method. In addition, it can be seen that on the Indonesian dataset, the performance of the second-order pseudo-perplexity is slightly higher than that of the first-order pseudo-perplexity, which is opposite to the phenomenon on the Tagalog dataset. Our method is less effective than the IndoGEC method in the two types of "conjunction" and "preposition," which are extremely confusing words, indicating that when the size of the confusing words reaches a certain number, the BERT model cannot distinguish between different words well. 

\section{Conclusion}
This study proposes a BERT-based unsupervised GEC framework, where GEC is viewed as multi-class classification task. We propose a scoring method for pseudo-perplexity to evaluate a sentence's probable correctness and constructs a Tagalog corpus for Tagalog GEC research. It obtains competitive performance on the Tagalog corpus we construct and open-source Indonesian corpus and it demonstrates that our framework is complementary to baseline method for low-resource GEC task. In the future, we will explore the feasibility of unsupervised learning in multilingual grammatical error correction tasks.
\section{Acknowledgments}

This document is the results of the National Social Science Fund of China (No. 22BTQ068), and the Science and Technology Program of Guangzhou (No.202002030227).

\bibliographystyle{ACM-Reference-Format}
\bibliography{sample-base}

\end{document}